\documentclass[conference]{IEEEtran}
\IEEEoverridecommandlockouts

\usepackage{cite}

\ifCLASSINFOpdf
   \usepackage[pdftex]{graphicx}
  \DeclareGraphicsExtensions{.pdf,.jpeg,.png}
\else
  \usepackage[dvips]{graphicx}
  \DeclareGraphicsExtensions{.eps}
\fi
\usepackage[cmex10]{amsmath}
\usepackage{amssymb}
\usepackage{dsfont}

\usepackage{mdwmath}
\usepackage{mdwtab}

\usepackage{pstricks,pst-plot,pst-node,setspace}
\usepackage{t1enc}
\usepackage{psfrag}

\usepackage[tight,footnotesize]{subfigure}
\usepackage{url}

\usepackage{comment}

\newcommand{\vect}[1]{ \boldsymbol{#1} }
\newcommand{\matr}[1]{ \boldsymbol{#1} }

\newcommand{\hvect}[1]{ \widehat{\boldsymbol{#1}} }

\renewcommand{\Im}{\text{Im}}

\newcommand{\CGaussPDF}{\mathrm{CN}}

\newcommand{\GammaPDF}{\mathrm{Ga}}
\newcommand{\ipilot}{\mathcal{P}}
\newcommand{\compnum}{\mathbb{C}}

\DeclareMathOperator*{\argmin}{argmin}

\DeclareMathOperator*{\diag}{diag}

\newcommand{\trans}{\mathrm{T}}
\newcommand{\hermit}{\mathrm{H}}

\begin{document}
%
\title{Application of Bayesian Hierarchical Prior Modeling to Sparse Channel Estimation}
\author{\IEEEauthorblockN{Niels Lovmand Pedersen\IEEEauthorrefmark{1}, 
Carles Navarro Manch\'{o}n\IEEEauthorrefmark{1}, 
Dmitriy Shutin\IEEEauthorrefmark{2}
and Bernard Henri Fleury\IEEEauthorrefmark{1}}
\IEEEauthorblockA{\IEEEauthorrefmark{1}Department of Electronic Systems, Aalborg University\\
Niels Jernes Vej 12, DK-9220 Aalborg, Denmark, Email: \{nlp,cnm,bfl\}@es.aau.dk} 
\IEEEauthorblockA{\IEEEauthorrefmark{2}Institute of Communications and Navigation, German Aerospace Center\\
Oberpfaffenhofen, D-82234 Wessling, Germany,  Email: dmitriy.shutin@dlr.de}
\thanks{\copyright\; 2012 IEEE. Personal use of this material is permitted. Permission from IEEE must be obtained for all other users, including reprinting/republishing this material for advertising or promotional purposes, creating new collective works for resale or redistribution to servers or lists, or reuse of any copyrighted components of this work in other works.}
}


\maketitle

\begin{abstract}
Existing methods for sparse channel estimation typically provide an estimate computed as the solution maximizing an objective function defined as the sum of the log-likelihood function and a penalization term proportional to the $\ell_1$-norm of the parameter of interest. However, other penalization terms have proven to have strong sparsity-inducing properties. In this work, we design pilot-assisted channel estimators for OFDM wireless receivers within the framework of sparse Bayesian learning by defining hierarchical Bayesian prior models that lead to sparsity-inducing penalization terms. The estimators result as an application of the variational message-passing algorithm on the factor graph representing the signal model extended with the hierarchical prior models. Numerical results demonstrate the superior performance of our channel estimators as compared to traditional and state-of-the-art sparse methods.

\end{abstract}


%
\IEEEpeerreviewmaketitle

\section{Introduction}
During the last few years the research on compressive sensing techniques and sparse signal representations \cite{BaraniukTutorial2007,Wakin2008} applied to channel estimation has received considerable attention, see e.g., \cite{Bajwa2010,Taubock2008,Berger2010,Huang2010,ShutinFleuryVBSAGEChannel}. The reason is that, typically, the impulse response of the wireless channel has a few dominant multipath components. 
A channel exhibiting this property is said to be sparse \cite{Bajwa2010}. 

The general goal of sparse signal representations from overcomplete dictionaries is to estimate the sparse vector $\vect{\alpha}$ in the following system model:
\begin{align}
\vect{y} = \matr{\Phi}\vect{\alpha} + \vect{w}.
\label{eq:sparsemodel}
\end{align}
In this expression $\vect{y} \in \compnum^M$ is the vector of measurement samples and $\vect{w} \in \compnum^M$ represents the samples of the additive white Gaussian random noise with covariance matrix $\lambda^{-1}\matr{I}$ and precision parameter $\lambda > 0 $. The matrix $\vect{\Phi}=[\vect{\phi}_1,\ldots,\vect{\phi}_L] \in \compnum^{M\times L}$ is the overcomplete dictionary with more columns than rows ($L > M$) and $\vect{\alpha} = [\alpha_1,\ldots,\alpha_L]^\trans \in \compnum^L$ is an unknown sparse vector, i.e., $\vect{\alpha}$ has few nonzero elements at unknown locations. 

Often, a sparse channel estimator is constructed by solving the $\ell_1$-norm constrained quadratic optimization problem, see among others \cite{Taubock2008,Berger2010,Huang2010}:
\begin{align}
\hvect{\alpha} = \argmin_{\vect{\alpha}} \left\{ \|\vect{y}-\matr{\Phi}\vect{\alpha}\|^2_2 + \kappa\|\vect{\alpha}\|_1 \right\}
\label{eq:lasso}
\end{align} 
with $\kappa > 0$ and $\|\cdot\|_p$, $p \geq 1$, denoting the $\ell_p$ vector norm. This method is also known as Least Absolute Shrinkage and Selection Operator (LASSO) regression \cite{Tibshirani1994} or Basis Pursuit Denoising \cite{Chen98atomicdecomposition}. The popularity of the LASSO regression is mainly attributed to the convexity of the cost function, as well as to its provable sparsity-inducing properties (see \cite{Wakin2008}). In \cite{Taubock2008,Berger2010,Huang2010} the LASSO regression is applied to \textit{orthogonal frequency-division multiplexing} (OFDM) pilot-assisted channel estimation. Various channel estimation algorithms that minimize the LASSO cost function using convex optimization are compared in \cite{Huang2010}.

Another approach to sparse channel estimation is sparse Bayesian learning (SBL) \cite{ShutinFleuryVBSAGEChannel,Tipping2001, WipfRao04,Tzikas2008}. 
Specifically, SBL aims at finding a sparse \emph{maximum a posteriori} (MAP) estimate of $\vect{\alpha}$
\begin{align}
\hvect{\alpha} = \argmin_{\vect{\alpha}} \left\{ \|\vect{y}-\matr{\Phi}\vect{\alpha}\|^2_2 + \lambda^{-1}Q(\vect{\alpha}) \right\}
\label{eq:map_general}
\end{align} 
by specifying a prior $p(\vect{\alpha})$ such that the penalty term $Q(\vect{\alpha}) \propto^{e} -\log p(\vect{\alpha})$ induces a sparse estimate $\hvect{\alpha}$.\footnote{Here $x\propto^{e}y$ denotes $\exp(x) = \exp(\upsilon) \exp(y)$, and thus $x = \upsilon + y$, for some arbitrary constant $\upsilon$. We will also make use of $x\propto y$ which denotes $x = \upsilon y$ for some positive constant $\upsilon$.} 


Obviously, by comparing \eqref{eq:lasso} and \eqref{eq:map_general} the SBL framework realizes the LASSO cost function by choosing the Laplace prior $p(\vect{\alpha}) \propto \exp(-a \|\vect{\alpha}\|_1)$ with $\kappa = \lambda^{-1}a$.
However, instead of working directly with the prior $p(\vect{\alpha})$, SBL models this using a two-layer (2-L) hierarchical structure. This involves specifying a conditional prior $p(\vect{\alpha}|\vect{\gamma})$ and a hyperprior $p(\vect{\gamma})$ such that
$
p(\vect{\alpha}) = \int p(\vect{\alpha}|\vect{\gamma})p(\vect{\gamma}) \textrm{d}\vect{\gamma}
$
has a sparsity-inducing nature. The hierarchical approach to the representation of $p(\vect{\alpha})$ has several important advantages. First of all, one is free to choose simple and analytically tractable probability density functions (pdfs). 
Second, when carefully chosen, the resulting hierarchical structure allows for the construction of efficient yet computationally tractable iterative inference algorithms with analytical derivation of the inference expressions. 

In \cite{Pedersen2011} we propose a 2-L and a three-layer (3-L) prior model for $\vect{\alpha}$. These hierarchical prior models lead to novel sparsity-inducing priors that  include the Laplace prior for complex variables as a special case. This paper adapts the Bayesian probabilistic framework introduced in \cite{Pedersen2011} to OFDM pilot-assisted sparse channel estimation. We then propose a variational message passing (VMP) algorithm that effectively exploits the hierarchical structure of the prior models. This approach leads to novel channel estimators that make use of various priors with strong sparsity-inducing properties. The numerical results reveal the promising potential of our estimators with improved performance as compared to state-of-the-art methods. In particular, the estimators outperform LASSO.

Throughout the paper we shall make use of the following notation: $(\cdot)^\trans$ and $(\cdot)^\hermit$ denote respectively the transpose and the Hermitian transpose; the expression $\langle f(\vect{x})\rangle_{q(\vect{x})}$ denotes the expectation of the function $f(\vect{x})$ with respect to the density $q(\vect{x})$;
$\CGaussPDF(\vect{x}|\vect{a},\matr{B})$ denotes a multivariate complex Gaussian pdf with mean $\vect{a}$ and covariance matrix $\matr{B}$; similarly, $\GammaPDF(x|a,b)= \frac{b^a}{\Gamma(a)}x^{a-1}\exp(-bx)$ denotes a Gamma pdf with shape parameter $a$ and rate parameter $b$.


\section{Signal Model}
We consider a single-input single-output OFDM system with $N$ subcarriers. A cyclic prefix (CP) is added to preserve orthogonality between subcarriers and to eliminate inter-symbol interference between consecutive OFDM symbols. The channel is assumed static during the transmission of each OFDM symbol. The received (baseband) OFDM signal $\vect{r} \in \compnum^N$ reads in matrix-vector notation
\begin{align}
	\vect{r} = \matr{X}\vect{h}+\matr{n}.
	\label{eq:ofdm}
\end{align}
The diagonal matrix $\matr{X} = \diag ( x_1, x_2, \ldots , x_N )$ contains the transmitted symbols. The components of the vector $\vect{h} \in \compnum^{N}$ are the samples of the channel frequency response at the $N$ subcarriers. Finally, $\vect{n} \in \compnum^N $ is a zero-mean complex symmetric Gaussian random vector of independent components with variance $\lambda^{-1}$.

To estimate the vector $\vect{h}$ in \eqref{eq:ofdm}, a total of $M$ pilot symbols are transmitted at selected subcarriers. The pilot pattern $\ipilot \subseteq \{1,\ldots,N\}$ denotes the set of indices of the pilot subcarriers. The received signals observed at the pilot positions $\vect{r}_\ipilot$ are then divided each by the corresponding pilot symbol $\matr{X}_\ipilot = \diag ( x_n : n \in \ipilot )$ to produce the vector of observations: 
\begin{align}
	\vect{y} \triangleq (\matr{X}_\ipilot)^{-1}\vect{r}_{\ipilot} = \vect{h}_{\ipilot} + (\matr{X}_\ipilot)^{-1}\vect{n}_{\ipilot}.
	\label{eq:pilotobs}
\end{align} 
We assume that all pilot symbols hold unit power such that the statistics of the noise term $(\matr{X}_\ipilot)^{-1}\vect{n}_\ipilot$ remain unchanged, i.e., $\vect{y} \in \compnum^{M}$ yields the samples of the true channel frequency response (at the pilot subcarriers) corrupted by additive complex white Gaussian noise with component variance $\lambda^{-1}$.

In this work, we consider a frequency-selective wireless channel that remains constant during the transmission of each OFDM symbol. The maximum relative delay $\tau_\textrm{max}$ is assumed to be large compared to the sampling time $T_\mathrm{s}$, i.e., $\tau_\textrm{max}/T_\mathrm{s} \gg 1$ \cite{Bajwa2010}. The impulse response of the wireless channel is modeled as a sum of multipath components:
\begin{align}
	g(\tau) = \sum_{k=1}^{K} \beta_k\delta\left(\tau - \tau_k\right).
	\label{eq:channel}
\end{align}
In this expression, $\beta_k$ and $\tau_k$ are respectively the complex weight and the continuous delay of the $k$th multipath component, and $\delta(\cdot)$ is the Dirac delta function. The parameter $K$ is the total number of multipath components. The channel parameters $K$, $\beta_k$, and $\tau_k$, $k=1,\ldots,K$, are random variables. Specifically, the weights $\beta_k$, $k=1,\ldots,K$, are mutually uncorrelated zero-mean with the sum of their variances normalized to one. 
Additional details regarding the assumptions on the model \eqref{eq:channel} are provided in Section \ref{sec:simulations}. 

\section{The Dictionary Matrix}
Our goal is to estimate $\vect{h}$ in \eqref{eq:ofdm} by applying the general optimization problem \eqref{eq:map_general} to the observation model \eqref{eq:pilotobs}. For doing so, we must define a proper dictionary matrix $\matr{\Phi}$. In this section we give an example of such a matrix. As a starting point, we invoke the parametric model \eqref{eq:channel} of the channel. Making use of this model, \eqref{eq:pilotobs} can be written as 
\begin{align}
	\vect{y} = \matr{T}(\vect{\tau})\vect{\beta} + \vect{w}
	\label{eq:yobs}
\end{align}
with $\vect{h}_\ipilot = \matr{T}(\vect{\tau})\vect{\beta}$, $\vect{w} = (\matr{X}_\ipilot)^{-1}\vect{n}_\ipilot$, $\vect{\beta} = \left[\beta_1,\ldots,\beta_K\right]^\trans$, $\vect{\tau} = \left[\tau_1,\ldots,\tau_K\right]^\trans$, and $\matr{T}(\vect{\tau}) \in \compnum^{M\times K}$ depending on the pilot pattern $\ipilot$ as well as the unknown delays in $\vect{\tau}$. Specifically, the $(m,k)$th entry of $\matr{T}(\vect{\tau})$ reads
\begin{align}
	\matr{T}(\vect{\tau})_{m,k} \triangleq \exp\left(-j2\pi f_m \tau_k\right), \begin{array}{l}  m = 1,2,\ldots,M \\ \;k = 1,2,\ldots,K \end{array}
	\label{eq:Measindices}
\end{align}
with $f_m$ denoting the frequency of the $m$th pilot subcarrier. In the general optimization problem \eqref{eq:map_general} the columns of $\matr{\Phi}$ are known. However, the columns of $\matr{T}(\vect{\tau})$ in \eqref{eq:yobs} depend on the unknown delays in $\vect{\tau}$. To circumvent this discrepancy we follow the same approach as in \cite{Berger2010} and consider a grid of uniformly-spaced delay samples in the interval $[0,\tau_{\mathrm{max}}]$: 
\begin{align}
	\vect{\tau}_d = \Big[0,\frac{T_\mathrm{s} }{\zeta},\frac{2T_\mathrm{s} }{\zeta}, \ldots, \tau_{\mathrm{max}} \Big]^\trans
\label{eq:delay_sampled}
\end{align}
with $\zeta>0$ such that $\zeta \tau_{\mathrm{max}} / T_\mathrm{s}$ is an integer. We now define the dictionary $\matr{\Phi} \in \compnum^{M\times L}$ as $\matr{\Phi} = \matr{T}(\vect{\tau}_d)$. Thus, the entries of $\matr{\Phi}$ are of the form \eqref{eq:Measindices} with delay vector $\vect{\tau}_d$. The number of columns $L = \zeta \tau_{\mathrm{max}} / T_\mathrm{s} + 1$ in $\matr{\Phi}$ is thereby inversely proportional to the selected delay resolution $T_\mathrm{s} /\zeta$. 

It is important to notice that the system model \eqref{eq:sparsemodel} with $\vect{\Phi}$ defined using discretized delay components is an approximation of the true system model \eqref{eq:yobs}. This approximation model is introduced so that \eqref{eq:map_general} can be applied to solve the channel estimation task. The estimate of the channel vector at the pilot subcarriers is then $\hvect{h}_\ipilot = \matr{\Phi}\hvect{\alpha}$. In order to estimate the channel $\vect{h}$ in \eqref{eq:ofdm} the dictionary $\vect{\Phi}$ is appropriately expanded (row-wise) to include all $N$ subcarrier frequencies. 

\section{Bayesian Prior Modeling}
In this section we specify the joint pdf of the system model \eqref{eq:sparsemodel} when it is augmented with the 2-L and the 3-L hierarchical prior model. The joint pdf of \eqref{eq:sparsemodel} augmented with the 2-L hierarchical prior model reads
\begin{align}
p(\vect{y},\vect{\alpha},\vect{\gamma},\lambda) = p(\vect{y}|\vect{\alpha},\lambda)p(\lambda)p(\vect{\alpha}|\vect{\gamma})p(\vect{\gamma};\vect{\eta}).
	\label{eq:jointpdf2layer}
\end{align}
The 3-L prior model considers the parameter $\vect{\eta}$ specifying the prior of $\vect{\gamma}$ in \eqref{eq:jointpdf2layer} as random. Thus, the joint pdf of \eqref{eq:sparsemodel} augmented with this hierarchical prior model is of the form
\begin{align}
p(\vect{y},\vect{\alpha},\vect{\gamma},\vect{\eta},\lambda) = p(\vect{y}|\vect{\alpha},\lambda)p(\lambda)p(\vect{\alpha}|\vect{\gamma})p(\vect{\gamma}|\vect{\eta})p(\vect{\eta}).
	\label{eq:jointpdf1}
\end{align}
In \eqref{eq:jointpdf2layer} and \eqref{eq:jointpdf1} we have $p(\vect{y}|\vect{\alpha},\lambda) = \CGaussPDF(\vect{y}|\matr{\Phi}\vect{\alpha},\lambda^{-1}\matr{I})$ due to \eqref{eq:sparsemodel}. Furthermore, we select the conjugate prior $p(\lambda) = p(\lambda; c,d) \triangleq \GammaPDF(\lambda|c,d)$. 
Finally, we let $p(\vect{\alpha}|\vect{\gamma}) = \prod_{l=1}^{L} p(\alpha_l|\gamma_l)$ with $p(\alpha_l|\gamma_l) \triangleq  \CGaussPDF(\alpha_l|0,\gamma_l)$. In the following we show the main results and properties of these prior models. We refer to \cite{Pedersen2011} for a more detailed analysis.


\subsection{Two-Layer Hierarchical Prior Model}
\begin{figure}[!t]
\centering
\subfigure[\label{fig:LogPenaltyL2}]{\includegraphics[width=0.42\linewidth]{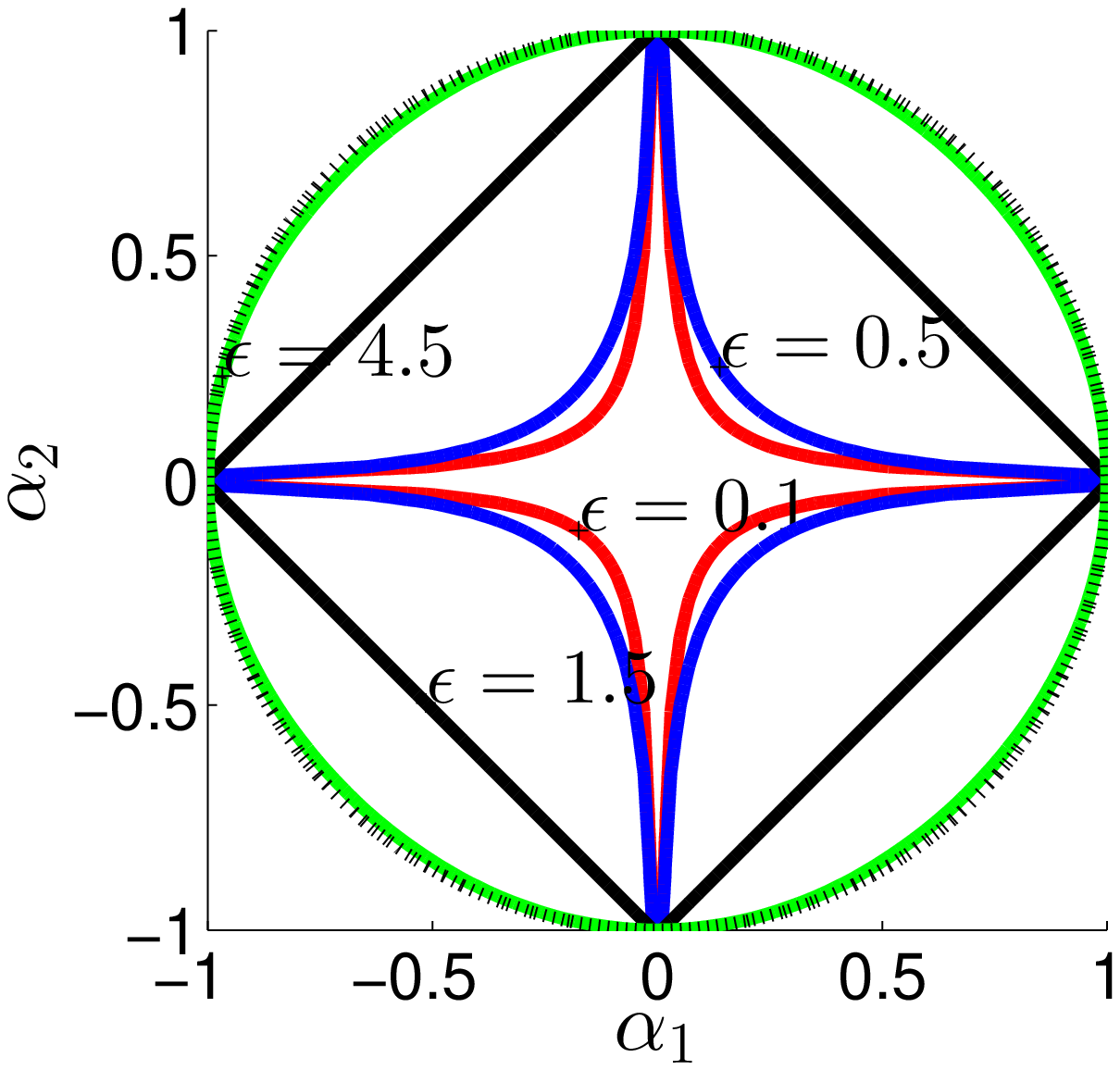}}
\subfigure[\label{fig:EstimationRulesL2}]{\includegraphics[width=0.51\linewidth]{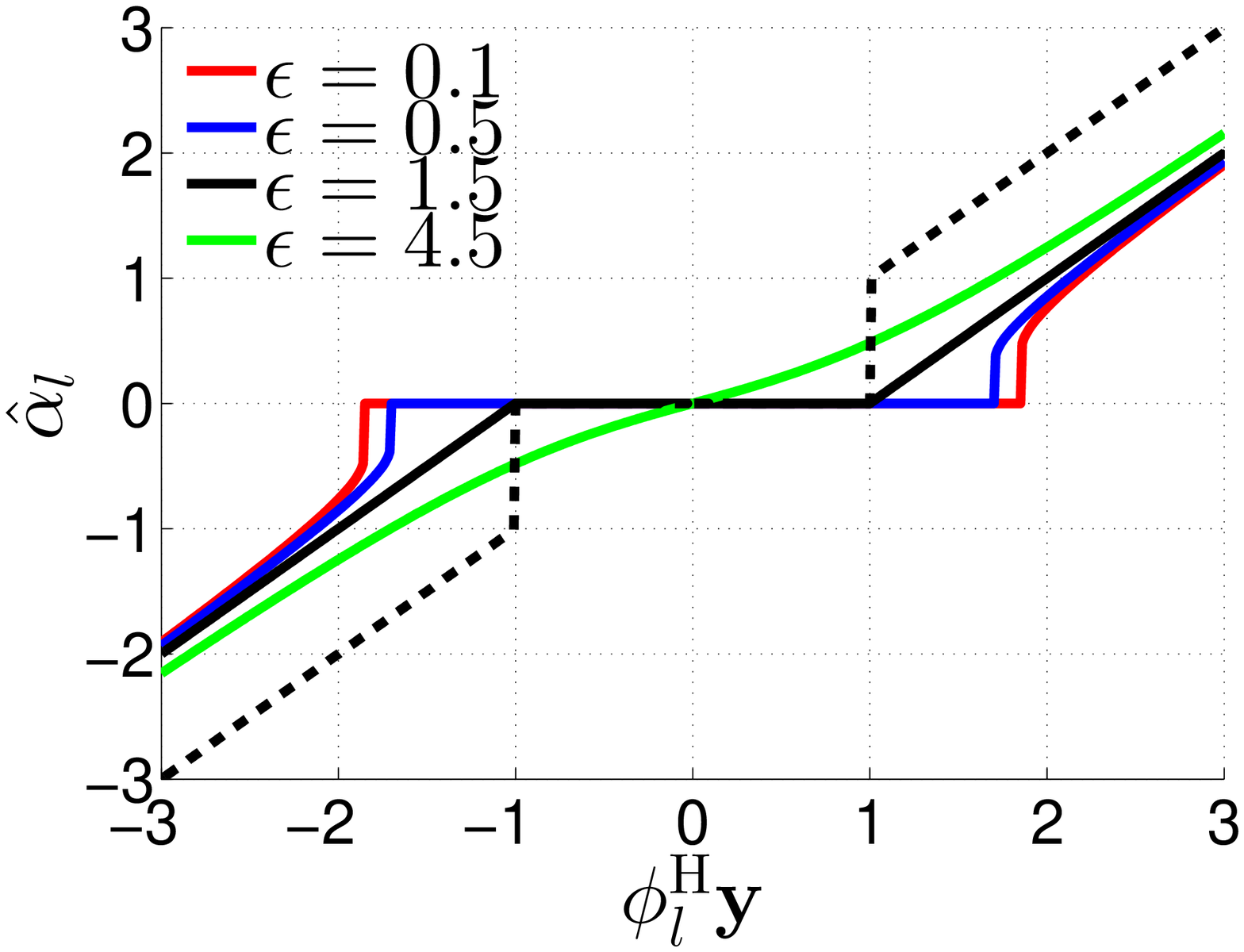}}
\caption{2-L hierarchical prior pdf for $\vect{\alpha} \in \compnum^2$: (a) Contour plot of the restriction to the $\Im\{\alpha_1\} = \Im\{\alpha_2\} = 0$ -- plane of the penalty term $Q(\alpha_1,\alpha_2;\epsilon,\eta) \propto^{e} -\log (p(\alpha_1;\epsilon,\eta)p(\alpha_2;\epsilon,\eta))$. (b) Restriction to $\Im\{\vect{\phi}^\hermit_l\vect{y}\} = 0$ of the resulting MAP estimation rule \eqref{eq:map_general} with $\epsilon$ as a parameter in the case when $\matr{\Phi}$ is orthonormal. The black dashed line indicates the hard-threshold rule and the black solid line the soft-threshold rule (obtained with $\epsilon =3/2$). The black dashed line indicates the penalty term resulting when the prior pdf is a circular symmetric Gaussian pdf. \label{fig:palphaL2}} 
\end{figure}
%

The 2-L prior model assumes that $p(\vect{\gamma})=\prod_{l=1}^L p(\gamma_l)$ with $p(\gamma_l) = p(\gamma_l;\epsilon,\eta_l) \triangleq \GammaPDF(\gamma_l|\epsilon,\eta_l)$. We compute the prior of $\vect{\alpha}$ to be 
\begin{align}
	p(\vect{\alpha};\epsilon,\vect{\eta})=\int_0^{\infty} p(\vect{\alpha}|\vect{\gamma})p(\vect{\gamma};\epsilon,\vect{\eta})\mathrm{d}\vect{\gamma}=\prod_{l=1}^L p(\alpha_l;\epsilon,\eta_l)
	\label{eq:pvectalpha}
\end{align}
with 
\begin{align}
p(\alpha_l;\epsilon,\eta_l)= \frac{2}{\pi \Gamma(\epsilon)}\eta_l^{\frac{(\epsilon+1)}{2}}|\alpha_l|^{\epsilon -1}K_{\epsilon-1}(2\sqrt{\eta_l}|\alpha_l|).
\label{eq:palpha}
\end{align}
In this expression, $K_\nu(\cdot)$ is the modified Bessel function of the second kind with order $\nu \in \mathds{R}$. The prior \eqref{eq:palpha} leads to the general optimization problem \eqref{eq:map_general} with penalty term
\begin{align}
Q(\vect{\alpha};\epsilon,\vect{\eta}) = \sum_{l=1}^{L}\log\left(|\alpha_l|^{\epsilon-1} K_{\epsilon-1}\left(2\sqrt{\eta_l}|\alpha_l|\right)\right).
\label{eq:weightedl2pm}
\end{align} 

We now show that the 2-L prior model induces the $\ell_1$-norm penalty term and thereby the LASSO cost function as a special case. Selecting $\epsilon=3/2$ and using the identity $K_{\frac{1}{2}}(z) = \sqrt{\frac{\pi}{2z}}\exp(-z)$ \cite{Abramowitz}, \eqref{eq:palpha} yields the Laplace prior
\begin{align}
p(\alpha_l; \epsilon = 3/2, \eta_l) = \frac{2\eta_l}{\pi}\exp(-2\sqrt{\eta_l}|\alpha_l|).
\label{eq:palphacomplex}
\end{align}
With the selection $\eta_l=\eta$, $l=1,\ldots,L$, we obtain $Q(\vect{\alpha};\eta) =  2\sqrt{\eta}\|\vect{\alpha}\|_1$.

The prior pdf \eqref{eq:palpha} is specified by $\epsilon$ and the regularization parameter $\vect{\eta}$. In order to get insight into the impact of $\epsilon$ on the properties of this prior pdf we consider the case $\vect{\alpha} \in \compnum^2$. In Fig.~\ref{fig:LogPenaltyL2} the contour lines of the restriction to $\mathbb{R}^2$ of $Q(\alpha_1,\alpha_2;\epsilon,\eta) \propto^{e} -\log (p(\alpha_1;\epsilon,\eta)p(\alpha_2;\epsilon,\eta))$ are visualized;\footnote{Let $f$ denote a function defined on a set $A$. The restriction of $f$ to a subset $B\subset A$ is the function defined on $B$ that coincides with $f$ on this subset.} each contour line is computed for a specific choice of $\epsilon$. Notice that as $\epsilon$ decreases towards $0$ more probability mass accumulates along the $\vect{\alpha}$-axes; as a consequence, the mode of the resulting posterior is more likely to be located close to the axes, thus promoting a sparse solution. The behavior of the classical $\ell_1$ penalty term obtained for $\epsilon=3/2$ can also be clearly recognized. In Fig.~\ref{fig:EstimationRulesL2} we consider the case when $\matr{\Phi}$ is orthonormal and compute the MAP estimator \eqref{eq:map_general} with penalty term \eqref{eq:weightedl2pm} for different values of $\epsilon$. Note the typical soft-threshold-like behavior of the estimators. As $\epsilon\rightarrow 0$, more components of $\hvect{\alpha}$ are pulled towards zero since the threshold value increases, thus encouraging a sparser solution. 

\subsection{Three-Layer Hierarchical Prior Model}
\begin{figure}[!t]
\centering
\subfigure[\label{fig:EstimationRulesL3}]{\includegraphics[width=0.51\linewidth]{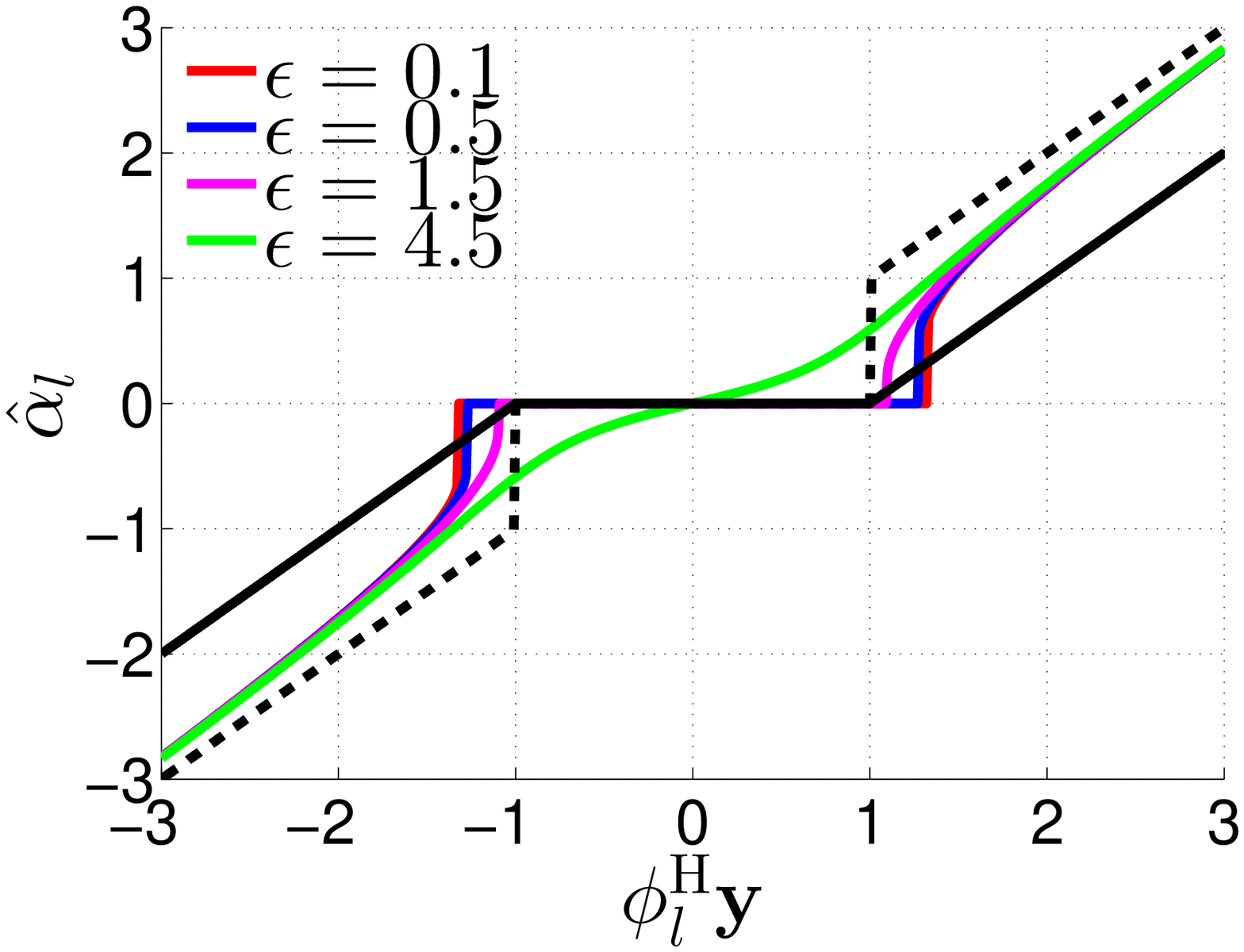}}
\subfigure[\label{fig:LogPenaltyL3}]{\includegraphics[width=0.42\linewidth]{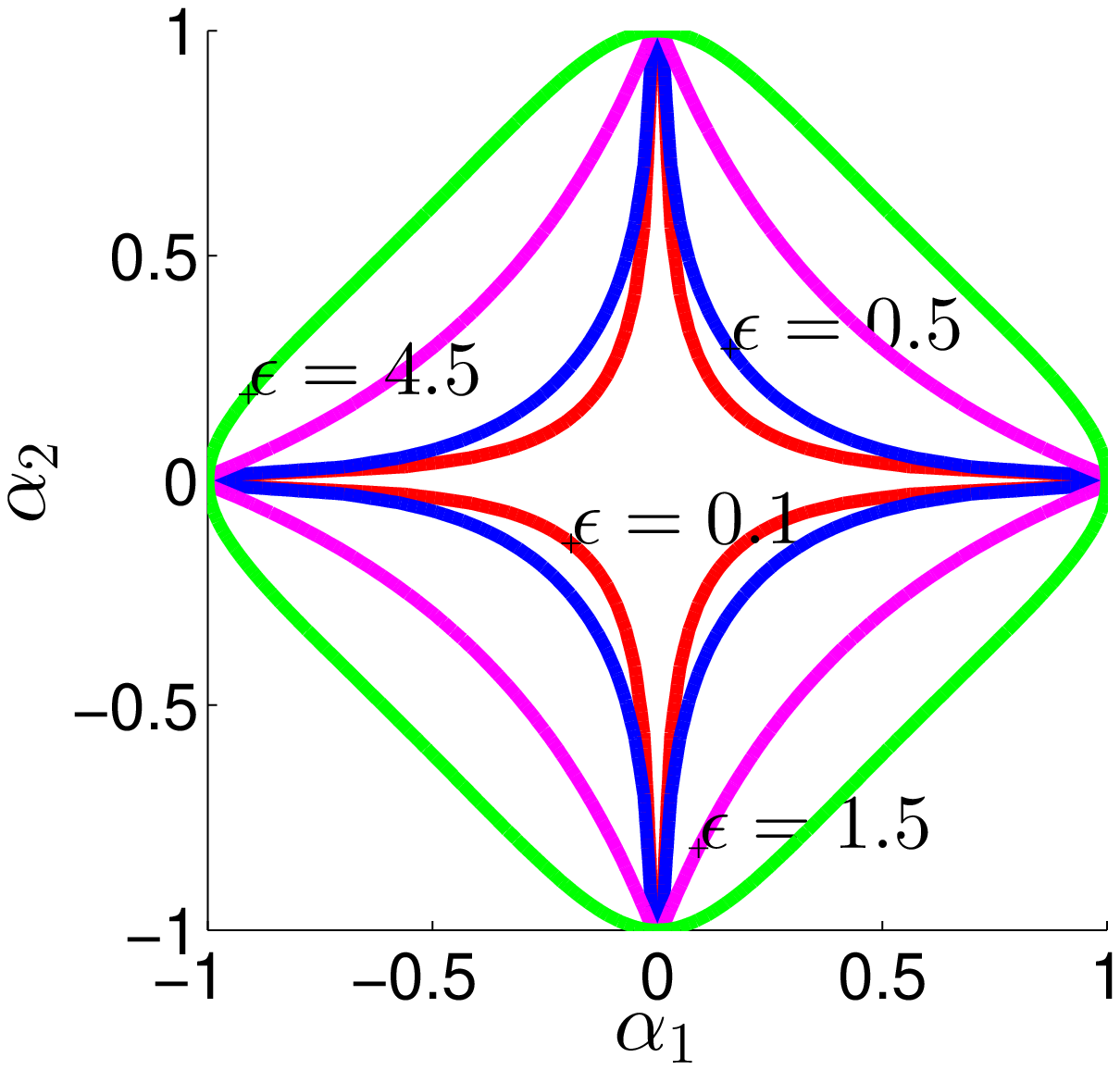}}
\caption{Three-layer hierarchical prior pdf for $\vect{\alpha} \in \compnum^2$ with the setting $a = 1$, $b = 0.1$: (a) Restriction to $\Im\{\vect{\phi}^\hermit_l\vect{y}\} = 0$ of the resulting MAP estimation rule \eqref{eq:map_general} with $\epsilon$ as a parameter in the case when $\matr{\Phi}$ is orthonormal. The black dashed line indicates the hard-threshold rule and the black solid line the soft-threshold rule. (b) Contour plot of the restriction to the $\Im\{\alpha_1\} = \Im\{\alpha_2\} = 0$ -- plane of the penalty term $Q(\alpha_1,\alpha_2;\epsilon,a,b) \propto^{e} - \log ( p(\alpha_1;\epsilon,a,b)p(\alpha_2;\epsilon,a,b))$. \label{fig:palphaL3}}
\end{figure} 

We now turn to the SBL problem with a 3-L prior model for $\vect{\alpha}$ leading to the joint pdf in \eqref{eq:jointpdf1}. Specifically, the goal is to incorporate the regularization parameter $\vect{\eta}$ into the inference framework. To that end, we define $p(\vect{\eta})=\prod^L_l p(\eta_l)$ with $p(\eta_l) =  p(\eta_l;a_l,b_l) \triangleq \GammaPDF(\eta_l|a_l,b_l)$ and compute the prior $p(\vect{\alpha})$. Defining  $\vect{a} \triangleq [a_1,\ldots,a_l]^\trans$ and $\vect{b} \triangleq [b_1,\ldots,b_L]^\trans$ we obtain  
$
p(\vect{\alpha};\epsilon,\vect{a},\vect{b})=\prod^L_l p(\alpha_l;\epsilon,a_l,b_l)
$ with
\begin{align}
	&p(\alpha_l;\epsilon,a_l,b_l)=\int_0^{\infty} p(\alpha_l|\gamma_l)p(\gamma_l)\mathrm{d}\gamma_l \notag \\
	&\quad = \frac{\Gamma(\epsilon+a_l)\Gamma(a_l+1)}{\pi b_l\Gamma(\epsilon)\Gamma (a_l)}\left(\frac{|\alpha_l|^2}{b_l}\right)^{\epsilon-1}
	U\left(\epsilon+a_l;\epsilon;\frac{|\alpha_l|^2}{b_l}\right).
\label{eq:palpha3D}
\end{align}   
In this expression, $U(\cdot;\cdot;\cdot)$ is the confluent hypergeometric function \cite{Abramowitz}. 
In Fig.~\ref{fig:EstimationRulesL3} we show the estimation rules produced by the MAP solver for different values of $\epsilon$ and fixed parameters $a_l$ and $b_l$ when $\matr{\Phi}$ is orthonormal. It can be seen that the estimation rules obtained with the 3-L prior model approximate the hard-thresholding rule. In Fig.~\ref{fig:LogPenaltyL3}, we depict the contour lines of the restriction to $\mathbb{R}^2$ of $Q(\alpha_1,\alpha_2;\epsilon,a,b) \propto^{e} - \log (p(\alpha_1;\epsilon,a,b)p(\alpha_2;\epsilon,a,b))$. Observe that although the contours behave qualitatively similarly to those shown in Fig. \ref{fig:LogPenaltyL2} for the 2-L prior model, the estimation rules in Fig.~\ref{fig:EstimationRulesL3} and Fig.~\ref{fig:EstimationRulesL2} are different.

Naturally, the 3-L prior model encompasses three free parameters, $\epsilon$, $\vect{a}$, and $\vect{b}$. The choice  $\epsilon=0$ and $b_l$ small (practically we let $b_l=10^{-6}$, $l=1,\ldots,L$) induces a weighted log-sum penalization term. This term is known to strongly promote a sparse estimate \cite{Tipping2001,WipfRao04}. Later in the text we will also adopt this parameter setting.

\section{Variational Message Passing \label{sec:vmp}}
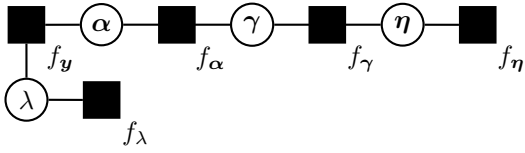
\begin{figure}[!t]
\centering
\scalebox{1}
{
\begin{pspicture}(0,0)(6.2,1.7)
		
    
    \cnodeput[fillcolor=white,fillstyle=solid](1.1,1.6){Alpha}{$\vect{\alpha}$}

    \cnodeput[fillcolor=white,fillstyle=solid](3.1,1.6){Gamma}{$\vect{\gamma}$}
    
    \cnodeput[fillcolor=white,fillstyle=solid](5.1,1.6){Eta}{$\vect{\eta}$}

    \cnodeput[fillcolor=white,fillstyle=solid](0.1,0.6){Lambda}{$\lambda$}

		\fnode*[framesize=0.5](0.1,1.6){Fy}
		\rput(0.55,1.15){\rnode[c]{py}{$f_{\vect{y}}$}} 
		
		\fnode*[framesize=0.5](2.1,1.6){Fa}
		\rput(2.55,1.15){\rnode[c]{pa}{$f_{\vect{\alpha}}$}}
		
		\fnode*[framesize=0.5](4.1,1.6){Fg}
		\rput(4.55,1.15){\rnode[c]{pg}{$f_{\vect{\gamma}}$}}
		
		\fnode*[framesize=0.5](6.1,1.6){Fe}
		\rput(6.55,1.15){\rnode[c]{pe}{$f_{\vect{\eta}}$}}		
				
		\fnode*[framesize=0.5](1.1,0.6){Fl}
		\rput(1.55,0.15){\rnode[c]{pl}{$f_{\lambda}$}}

    \ncline{-}{Alpha}{Fy}
    \ncline{-}{Gamma}{Alpha}
    \ncline{-}{Eta}{Gamma}
    \ncline{-}{Eta}{Fe}
    \ncline{-}{Lambda}{Fy}
    \ncline{-}{Fl}{Lambda}
    
\end{pspicture}
}
\caption{A factor graph that represents the joint pdf \eqref{eq:jointpdf1}. In this figure $f_{\vect{y}} \equiv p(\vect{y}|\vect{\alpha},\lambda)$, $f_{\vect{\alpha}} \equiv p(\vect{\alpha}|\vect{\gamma})$, $f_{\vect{\gamma}} \equiv p(\vect{\gamma}|\vect{\eta})$, $f_{\vect{\eta}} \equiv p(\vect{\eta})$, and $f_{\lambda} \equiv p(\lambda)$.}
\label{fig:fg_complex}
\end{figure}

In this section we present a VMP algorithm for estimating $\vect{h}$ in \eqref{eq:ofdm} given the observation $\vect{y}$ in \eqref{eq:pilotobs}. Let $\vect{\Theta} = \left\{\vect{\alpha},\vect{\gamma},\vect{\eta}, \lambda \right\}$ be the set of unknown parameters and $p(\vect{y},\vect{\Theta})$ be the joint pdf specified in \eqref{eq:jointpdf1}. The factor graph \cite{Kschischang} that encodes the factorization of $p(\vect{y},\vect{\Theta})$ is shown in Fig.~\ref{fig:fg_complex}. Consider an auxiliary pdf $q(\vect{\Theta})$ for the unknown parameters that factorizes according to $q(\vect{\Theta}) = q(\vect{\alpha})q(\vect{\gamma})q(\vect{\eta})q(\lambda)$.
The VMP algorithm is an iterative scheme that attempts to compute the auxiliary pdf that minimizes the Kullback-Leibler (KL) divergence $\text{KL}(q(\vect{\Theta})\|p(\vect{\Theta}|\vect{y}))$. In the following we summarize the key steps of the algorithm; the reader is referred to \cite{Winn} for more information on VMP. 

From \cite{Winn} the auxiliary function $q(\vect{\theta}_i)$, $\vect{\theta}_i \in \vect{\Theta}$, is updated as the product of incoming messages from the neighboring factor nodes $f_n$ to the variable node $\vect{\theta}_i$:
\begin{align}
\label{eq:vmp1}
q(\vect{\theta}_i) \propto \prod_{f_n \in \mathcal{N}_{\vect{\theta}_i}} m_{f_n\rightarrow\vect{\theta}_i}.
\end{align}
In \eqref{eq:vmp1} $\mathcal{N}_{\vect{\theta}_i}$ is the set of factor nodes neighboring the variable node $\vect{\theta}_i$ and $m_{f_n \rightarrow \vect{\theta}_i}$ denotes the message from factor node $f_n$ to variable node $\vect{\theta}_i$. This message is computed as
\begin{align}
\label{eq:vmp2}
m_{f_n\rightarrow \vect{\theta}_i} = \exp\left( \langle\ln f_n\rangle_{\prod_j q(\vect{\theta}_j),\; \vect{\theta}_j \in \mathcal{N}_{f_n}\backslash \{\vect{\theta}_i\}}\right),
\end{align}
where $\mathcal{N}_{f_n}$ is the set of variable nodes neighboring the factor node $f_n$.
After an initialization procedure, the individual factors of $q(\vect{\Theta})$ are then updated iteratively in a round-robin fashion using \eqref{eq:vmp1} and \eqref{eq:vmp2}.

We provide two versions of the VMP algorithm: one applied to the 2-L prior model (referred to as VMP-2L) and another one applied to the 3-L model (VMP-3L). The messages corresponding to VMP-2L are easily obtained as a special case of the messages computed for VMP-3L by assuming $q(\eta_l)=\delta(\eta_l-\hat{\eta_l})$, where $\hat{\eta_l}$ is some fixed real number. 

\subsubsection{Update of $q(\vect{\alpha})$}
According to \eqref{eq:vmp1} and Fig.~\ref{fig:fg_complex} the computation of the update of $q(\vect{\alpha})$ requires evaluating the product of messages $m_{f_{\vect{y}}\rightarrow \vect{\alpha}}$ and $m_{f_{\vect{\alpha}}\rightarrow \vect{\alpha}}$. Multiplying these two messages yields the Gaussian auxiliary pdf $q(\vect{\alpha}) = \mathrm{CN}\left( \vect{\alpha}|\hat{\vect{\alpha}}, \hat{\matr{\Sigma}}_{\vect{\alpha}} \right)$ with covariance matrix and mean given by
\begin{align}
\hat{\matr{\Sigma}}_{\vect{\alpha}} &= (\langle \lambda \rangle_{q(\lambda)}\matr{\Phi}^\hermit\matr{\Phi}+ \matr{V}(\vect{\gamma}))^{-1}, \label{eq:alpha_cov}\\
\hat{\vect{\alpha}} &= \langle \vect{\alpha} \rangle_{q(\vect{\alpha})} = \langle \lambda \rangle_{q(\lambda)}\hat{\matr{\Sigma}}_{\vect{\alpha}}\matr{\Phi}^\hermit\vect{y}. \label{eq:alpha_mean}
\end{align}
In the above expression we have defined $\matr{V}(\vect{\gamma}) = \diag(\langle\gamma^{-1}_1\rangle_{q(\vect{\gamma})},\ldots, \langle\gamma^{-1}_L\rangle_{q(\vect{\gamma})} )$.



\begin{figure*}[!t]
\centering
\centerline{
\subfigure[\label{fig:CodedBERvsEbN0}]{\includegraphics[width=0.32\linewidth]{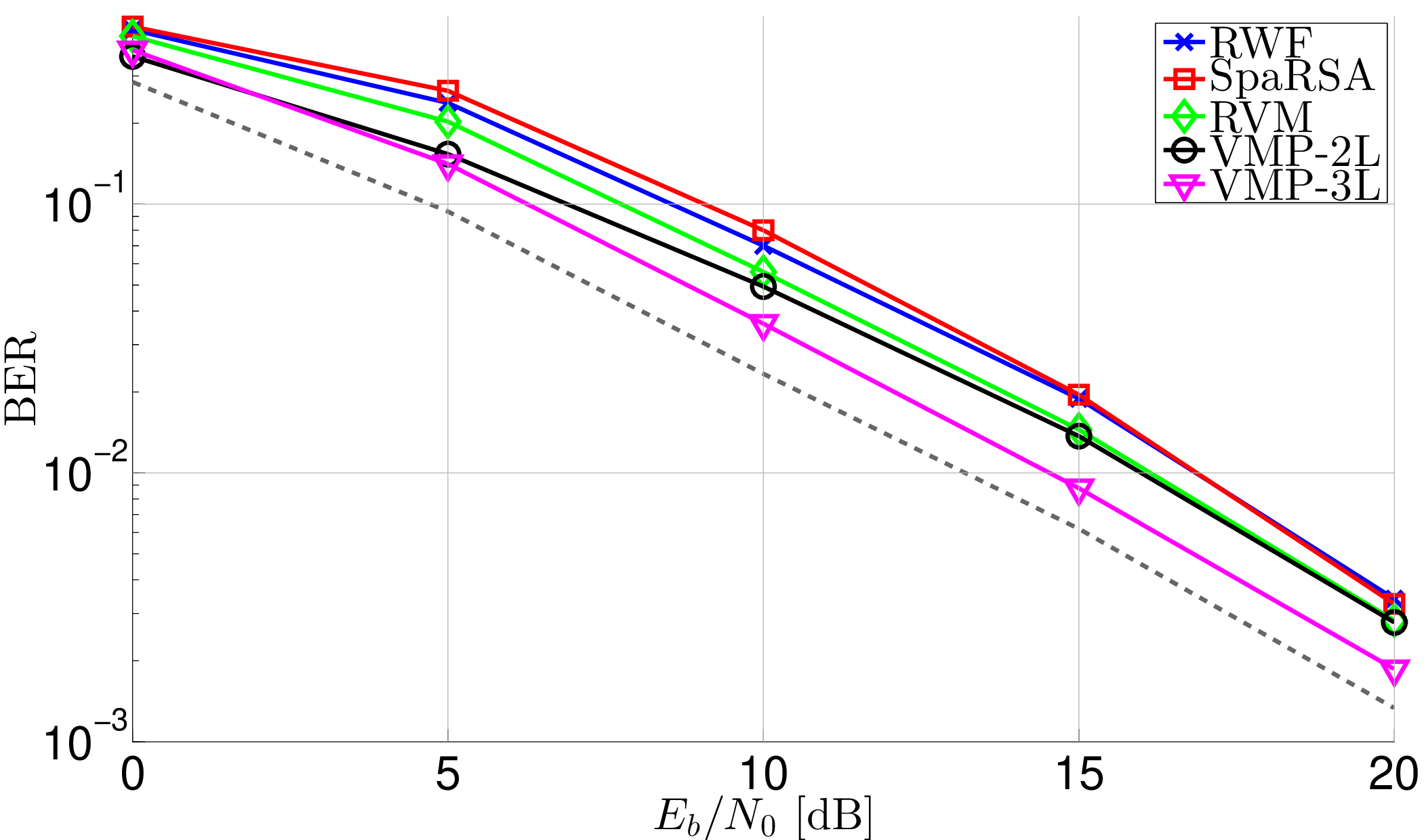}}
\subfigure[\label{fig:MSE_cod_vsEbN0}]{\includegraphics[width=0.32\linewidth]{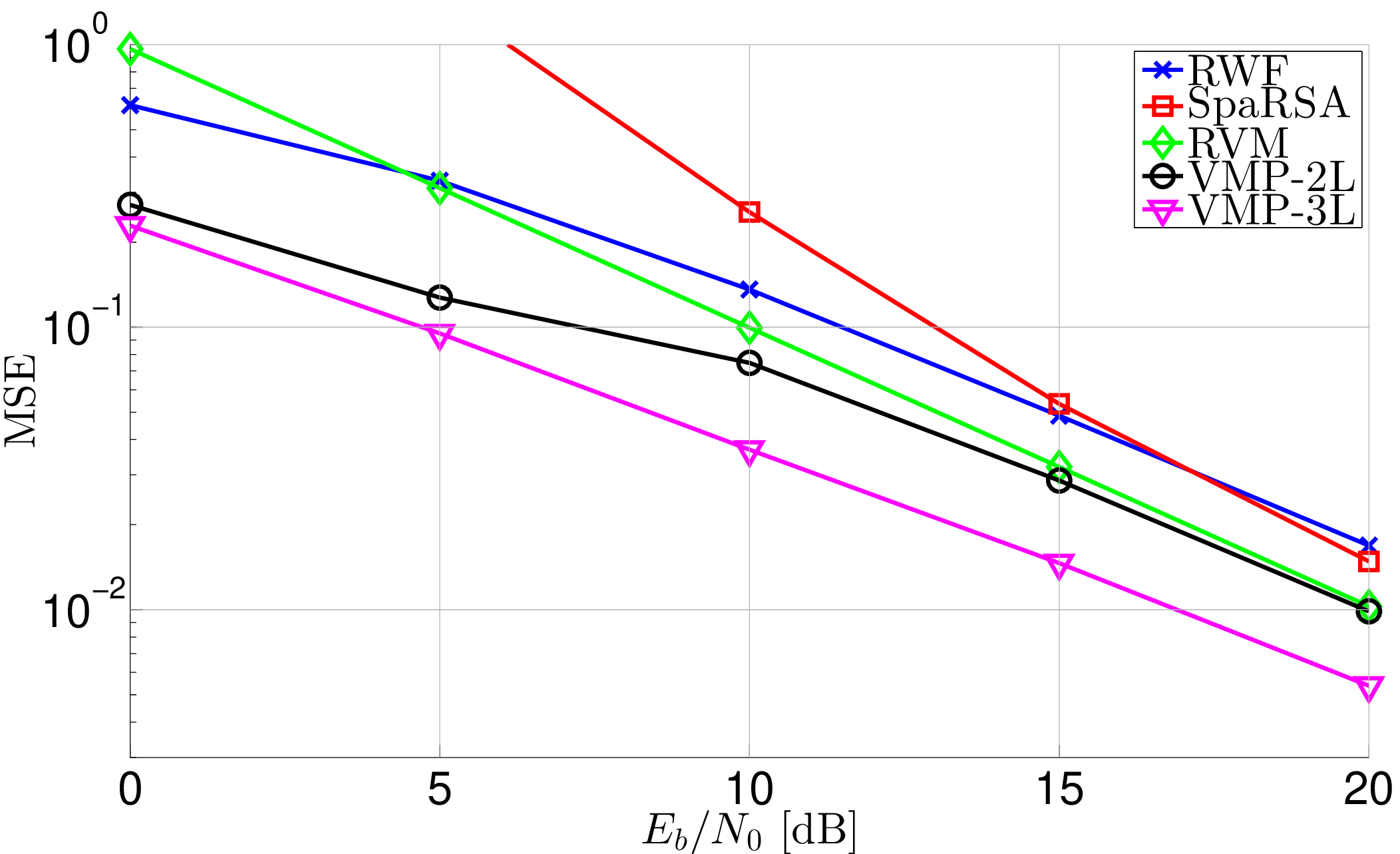}}
\subfigure[\label{fig:MSEvsNoPilots15}]{\includegraphics[width=0.32\linewidth]{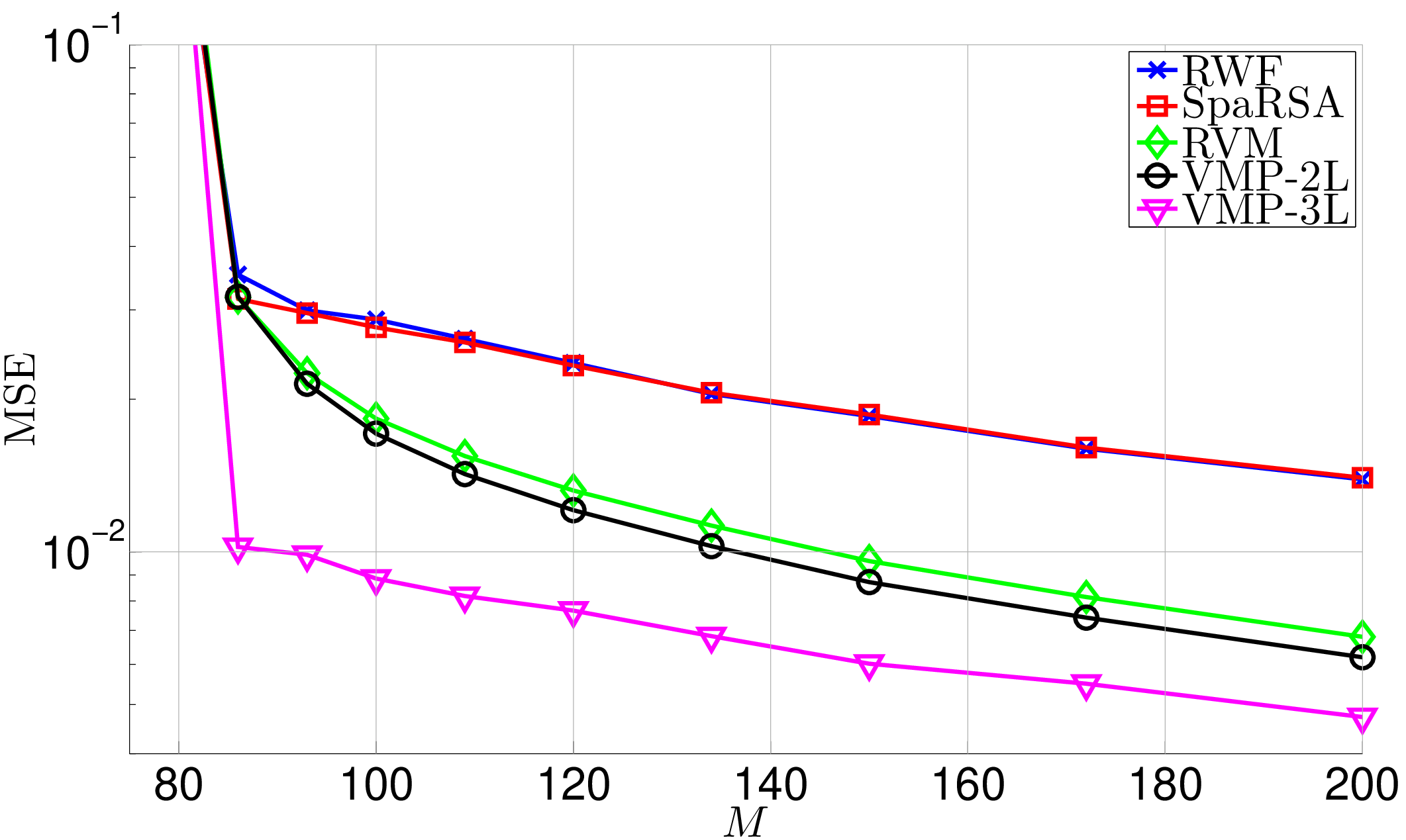}}
}
\caption{Comparison of the performance of the VMP-2L, VMP-3L, RWF, RVM, and SparseRSA algorithms: 
(a) BER versus $E_b/N_0$, (b) MSE versus $E_b/N_0$, (c) MSE versus number of available pilots $M$ with fixed $L = 200$ and the ratio between received symbol power and noise variance set to 15 dB. In (a,b) we have $M = 100$ and $L = 200$. In (a) the dashed line shows the BER performance when the true channel vector $\vect{h}$ in \eqref{eq:ofdm} is known.}
\label{fig:performance}
\end{figure*}

\subsubsection{Update of $q(\vect{\gamma})$}
The update of $q(\vect{\gamma})$ is proportional to the product of the messages $m_{ f_{\vect{\alpha} } \rightarrow \vect{\gamma} }$ and $m_{ f_{\vect{\gamma} } \rightarrow \vect{\gamma} }$:
\begin{align}
\label{eq:qg}
q(\vect{\gamma}) \propto \prod_{l=1}^{L} \gamma_l^{\epsilon-2} \exp \left( -\gamma_l^{-1}\langle|\alpha_l|^2\rangle_{q(\vect{\alpha})} - \gamma_l\langle\eta_l\rangle_{q(\vect{\eta})}  \right).
\end{align}
The right-hand side expression in \eqref{eq:qg} is recognized as the product of Generalized Inverse Gaussian (GIG) pdfs \cite{Joergensen} with order $p = \epsilon -1$. Observe that the computation of $\matr{V}(\vect{\gamma})$ in \eqref{eq:alpha_cov} requires evaluating $\langle\gamma_l^{-1}\rangle_{q(\vect{\gamma})}$ for all $l=1,\ldots,L$.
Luckily, the moments of the GIG distribution are given in closed form for any $n\in \mathbb{R}$ \cite{Joergensen}:
\begin{align}
\langle\gamma_l^n\rangle_{q(\vect{\gamma})} = \left(\frac{\langle|\alpha_l|^2\rangle_{q(\vect{\alpha})}}{\langle\eta_l\rangle_{q(\vect{\eta})}}\right)^{\frac{n}{2}} \frac{K_{p+n}\big(2\sqrt{\langle\eta_l\rangle_{q(\vect{\eta})}\langle|\alpha_l|^2\rangle_{q(\vect{\alpha})}}\big) }{K_{p}\big( 2\sqrt{\langle\eta_l\rangle_{q(\vect{\eta})}\langle|\alpha_l|^2\rangle_{q(\vect{\alpha})}}\big)} .
\label{eq:gamma_mean1}
\end{align}

\subsubsection{Update of $q(\vect{\eta})$}
The update of $q(\vect{\eta})$ is proportional to the product of messages $m_{ f_{\vect{\eta} } \rightarrow \vect{\eta} }$ and $m_{ f_{\vect{\gamma} } \rightarrow \vect{\eta} }$:
\begin{align}
q(\vect{\eta}) \propto  \prod_{l=1}^L \eta_l^{\epsilon+a_l-1}\exp\left(-( \langle \gamma_l \rangle_{q(\vect{\gamma})}+b_l )\eta_l\right).
\end{align}
Clearly, $q(\vect{\eta})$ factorizes as a product of $L$ gamma pdfs, one for each individual entry in $\vect{\eta}$. The first moment of $\eta_l$ used in \eqref{eq:gamma_mean1} is easily computed as
\begin{align}
\langle \eta_l \rangle_{q(\vect{\eta})} = \frac{\epsilon+a_l}{\langle \gamma_l \rangle_{q(\vect{\gamma})}+b_l }.
\label{eq:eta_mean}
\end{align}
Naturally, $q(\vect{\eta})$ is only computed for VMP-3L.

\subsubsection{Update of $q(\lambda)$}
It can be shown that $q(\lambda) = \mathrm{Ga}(\lambda | M+c,\langle\|\vect{y} - \matr{\Phi}\vect{\alpha}\|^2_2\rangle_{q(\vect{\alpha})}+d)$.
The first moment of $\lambda$ used in \eqref{eq:alpha_cov} and \eqref{eq:alpha_mean} is therefore
\begin{align}
\langle \lambda \rangle_{q(\lambda)} = \frac{ M+c}{ \langle\|\vect{y} - \matr{\Phi}\vect{\alpha}\|^2_2\rangle_{q(\vect{\alpha})} +d }.
\label{eq:lambda_mean}
\end{align}

\section{Numerical Results \label{sec:simulations}}
\begin{table}[!t]
	\centering
	\caption{Parameter settings for the simulations. The convolutional code and decoder has been implemented using \cite{cml}. \label{tab:ofdmsettings}}
	\vspace{-7pt} 
	\begin{tabular}{c|c}
	\hline
	Sampling time, $T_\mathrm{s}$ & 32.55 ns \\ 
	CP length & 4.69 $\mu$s / 144 $T_\mathrm{s}$\\
  Subcarrier spacing   & 15 kHz\\
	Pilot pattern & Equally spaced, QPSK\\
	Modulation & QPSK \\
	Subcarriers, $N$ & 1200 \\
	Pilots, $M$ & 100 \\
	OFDM symbols & 1 \\
	Information bits & 727 \\
	Channel interleaver & Random \\
	Convolutional code & $(133,171,165)_8$ \\
	Decoder & BCJR algorithm \cite{Bahl1974}
	\end{tabular}
\end{table}

We perform Monte Carlo simulations to evaluate the performance of the two versions of the derived VMP algorithm in Section~\ref{sec:vmp}. We consider a scenario inspired by the 3GPP LTE standard \cite{3GPP2008} with the settings specified in Table~\ref{tab:ofdmsettings}. The multipath channel \eqref{eq:channel} is based on the model used in \cite{Jakobsen2010} where, for each realization of the channel, the total number of multipath components $K$ is Poisson distributed with mean of $\langle K\rangle_{p(K)} = 10$ and the delays $\tau_k$, $k =1,\ldots,K$, are independent and uniformly distributed random variables drawn from the continuous interval [0, 144 $T_\mathrm{s}$] (corresponding to  the CP length). The $k$th nonzero component $\beta_k$ conditioned on the delay $\tau_k$ has a zero-mean complex circular symmetric Gaussian distribution with variance $\sigma^2(\tau_k) = \langle |\beta_k|^2 \rangle_{p(\beta_k|\tau_k)} = u\exp(-\tau_k/v)$ and parameters $u,v > 0$.\footnote{The parameter $u$ is computed such that $\langle \sum_{k=1}^{K} |\beta_k(t)|^2 \rangle_{p(\vect{\beta},\vect{\tau},K)} = 1$, where $p(\vect{\beta},\vect{\tau},K)$ is the joint pdf of the parameters of the channel model. In the considered simulation scenario, $\langle K\rangle_{p(K)} = 10$, $\tau_\textrm{max} = 144\; T_\mathrm{s}$, and $v = 20\; T_\mathrm{s}$ (the decay rate).}

To initialize the VMP algorithm we set $\langle \lambda \rangle_{q(\lambda)}$ and $\langle\gamma_l^{-1}\rangle_{q(\vect{\gamma})}$ equal to the inverse of the sample variance of $\vect{y}$ and the inverse number of columns $L$ respectively. Furthermore, we let $c = d = 0$ in \eqref{eq:lambda_mean}, which corresponds to the Jeffreys noninformative prior for $\lambda$. Once the initialization is completed, the algorithm sequentially updates the auxiliary pdfs $q(\vect{\alpha})$, $q(\vect{\gamma})$, $q(\vect{\eta})$, and $q(\lambda)$ until convergence is achieved. 
Obviously, $q(\vect{\eta})$ is only updated for VMP-3L, whereas for VMP-2L the entries in $\vect{\eta}$ are set to $M$. For both versions we select $\epsilon =0$ and for VMP-3L we set $a_l=1$ and $b_l=10^{-6}$, $l=1,\ldots,L$. Finally, the dictionary $\vect{\Phi}$ is specified by $M$ pilot subcarriers and a total of $L = 200$ columns (corresponding to the choice $\tau_{\mathrm{max}} = 144$ $T_\mathrm{s}$ and $\zeta \approx 1.4$ in \eqref{eq:delay_sampled}).

The VMP is compared to a classical OFDM channel estimator and two state-of-the-art sparse estimation schemes. Specifically, we use as benchmark the robustly-designed Wiener Filter (RWF) \cite{Edfors1998}, the relevance vector machine (RVM) \cite{Tipping2001}, \cite{WipfRao04},\footnote{The software is available on-line at \url{http://dsp.ucsd.edu/~dwipf/}.} and the \textit{sparse reconstruction by separable approximation} (SpaRSA) algorithm \cite{Wright2009}.\footnote{The software is available on-line at \url{http://www.lx.it.pt/~mtf/SpaRSA/}} The RVM is an EM algorithm based on the 2-L prior model of the student-t pdf over each $\alpha_l$, whereas SpaRSA is a proximal gradient method for solving \eqref{eq:lasso}. In case of the SpaRSA algorithm the regularization parameter $\kappa$ needs to be set. In all simulations, we let $\kappa = 2$, which leads to good performance in high signal-to-noise ratio (SNR) regime. 

The performance is compared with respect to the resulting bit-error-rate (BER) and mean-squared error (MSE) of the estimate $\hvect{h}$ versus the SNR ($E_b/N_0$). In addition, in order to quantify the necessary pilot overhead, we evaluate the MSE versus the number of available pilots $M$. Hence, in this setup $M$ is no longer fixed as in Table~\ref{tab:ofdmsettings}. 

In Fig.~\ref{fig:CodedBERvsEbN0} we compare the BER performance of the different schemes. We see that VMP-3L outperforms the other schemes across all the SNR range considered. Specifically, at 1 \% BER the gain is approximately 2 dB compared to VMP-2L and RVM and 3 dB compared to SpaRSA and RWF. Also VMP-2L achieves lower BER in the SNR range 0 - 12 dB compared to RVM and across the whole SNR range compared to SpaRSA and RWF. 

The superior BER performance of the VMP algorithm is well reflected in the MSE performance shown in Fig.~\ref{fig:MSE_cod_vsEbN0}. Again VMP-3L is a clear winner followed by VMP-2L. The bad MSE performance of the SpaRSA for low SNR is due to the difficulty in specifying a suitable regularization parameter $\kappa$ across a large SNR range. 

We next fix the ratio between received symbol power and noise variance to 15 dB\footnote{Note that this value does not correspond with $E_b/N_0$ as represented in Fig.~\ref{fig:CodedBERvsEbN0} and \ref{fig:MSE_cod_vsEbN0}. The specific $E_b/N_0$ depends on the number of bits in an OFDM block, which in turn depends on the number of pilot symbols $M$.} and evaluate the MSE versus number of available pilots $M$. The results are depicted in Fig.~\ref{fig:MSEvsNoPilots15}. Observe a noticeable performance gain obtained with VMP-3L. In particular, VMP-3L exhibits the same MSE performance as VMP-2L and RVM using only approximately 85 pilots, roughly half as many as VMP-2L and RVM. 
Furthermore, VMP-3L, using this number of pilots, significantly outperforms SpaRSA and RWF using 200 pilots.

\section{Conclusion}
In this paper, we proposed channel estimators based on sparse Bayesian learning. The estimators rely on Bayesian hierarchical prior modeling and variational message passing (VMP). The VMP algorithm effectively exploits the probabilistic structure of the hierarchical prior models and the resulting sparsity-inducing priors. Our numerical results show that the proposed channel estimators yield superior performance in terms of bit-error-rate and mean-squared error as compared to other existing estimators, including the estimator based on the $\ell_1$-norm constraint. They also allow for a significant reduction of the amount of pilot subcarriers needed for estimating a given channel.  
\section*{Acknowledgment}

This work was supported in part by the 4GMCT cooperative research project funded by Intel Mobile Communications, Agilent Technologies, Aalborg University and the Danish National Advanced Technology Foundation. This research was also supported in part by the project ICT- 248894 Wireless Hybrid Enhanced Mobile Radio Estimators (WHERE2). 

\IEEEtriggeratref{4}


%

\bibliographystyle{IEEEtran}
\bibliography{Hcomplex}

\end{document}